\documentclass[sigconf]{acmart}
\graphicspath{{figs/}}
\usepackage{multirow}
\usepackage{graphicx} %插入图片的宏包
\usepackage{float}
\usepackage{subfigure} %插入多图时用子图显示的宏包

\AtBeginDocument{%
  \providecommand\BibTeX{{%
    \normalfont B\kern-0.5em{\scshape i\kern-0.25em b}\kern-0.8em\TeX}}}

\begin{document}

\title{Discovering Traveling Companions using Autoencoders}


% \author{Anonymous}
% %\authornote{Both authors contributed equally to this research.}
% \email{Anonymous@anonymous}
% %\orcid{1234-5678-9012}
% %\author{G.K.M. Tobin}
% %\authornotemark[1]
% %\email{webmaster@marysville-ohio.com}
% \affiliation{%
%   \institution{Anonymous Institute}
%   %\streetaddress{P.O. Box 1212}
%   %\city{Dublin}
%   %\state{Ohio}
%   %\postcode{43017-6221}
% }

\author{Xiaochang Li}
\affiliation{%
  \institution{East China Normal University}}
\email{xcli@stu.ecnu.edu.cn}

\author{Bei Chen}
\affiliation{%
  \institution{East China Normal University}}
\email{51195100002@stu.ecnu.edu.cn}

\author{Xuesong Lu}
\affiliation{%
  \institution{East China Normal University}}
\email{xuesong.lu.dase@gmail.com}


\begin{abstract}
With the wide adoption of mobile devices, today's location tracking systems such as satellites, cellular base stations and wireless access points are continuously producing tremendous amounts of location data of moving objects. The ability to discover moving objects that travel together, i.e., traveling companions, from their trajectories is desired by many applications such as intelligent transportation systems and location-based services. Existing algorithms are either based on pattern mining methods that define a particular pattern of traveling companions or based on representation learning methods that learn similar representations for similar trajectories. The former methods suffer from the pairwise point-matching problem and the latter often ignore the temporal proximity between trajectories. In this work, we propose a generic deep representation learning model using autoencoders, namely, ATTN-MEAN, for the discovery of traveling companions. ATTN-MEAN collectively injects spatial and temporal information into its input embeddings using skip-gram, positional encoding techniques, respectively. Besides, our model further encourages trajectories to learn from their neighbours by leveraging the Sort-Tile-Recursive algorithm, mean operation and global attention mechanism. After obtaining the representations from the encoders, we run DBSCAN to cluster the representations to find travelling companion. The corresponding trajectories in the same cluster are considered as traveling companions. 
Experimental results suggest that ATTN-MEAN performs better than the state-of-the-art algorithms on finding traveling companions.
\end{abstract}





\maketitle
\section{Introduction}
Discovering moving objects that travel together, i.e., traveling companions, is an interesting problem in many real-world applications. For example, in mobile advertising it is proven that consumers who occurred
% are presented ,zubcsek2017predicting
at the same location at the same time generally exhibit commonalities in their taste~\cite{ghose2019mobile}. Thus detecting groups of consumers who walk together and advertising to them using coupons in the same product category might increase the advertising response. Other applications concerning with finding traveling companions include intelligent transportation systems, animal migration monitoring and public procession management.

Traveling companions can be discovered from the trajectories of moving objects. An object's trajectory is a sequence of its location points sorted chronologically. Traveling companions are thus moving objects with trajectories that are both spatially and temporally close. Existing methods for mining traveling companions can be broadly divided into two categories: methods based on pattern mining~\cite{jeung2008discovery,li2010swarm,zheng2013online} and methods based on representation learning~\cite{yao2017trajectory,li2018deep,yao2019computing}. Pattern mining based methods often first define a particular movement pattern pertaining to traveling companions based on some similarity measurement of their trajectories, and then develop specific algorithms to extract the predefined patterns. The similarity measurements are often based on point-wise Euclidean distance and thereby require trajectories are aligned along timestamp. However, real trajectories often contain many missing location points and have to be interpolated, which introduces additional measurement errors. Representation learning based methods, on the other hand, do not rely on pattern definition or point-wise comparison. Instead, they learn the representations of the trajectories using machine/deep learning models and then cluster the representations to discover similar ones, which in turn represent similar trajectories. Existing models either require certain amount of feature engineering, or specific labels for supervised learning. But in practice, feature engineering is often problem dependent and time consuming, which introduces extra overhead to representation learning itself. Yet trajectory labels are usually unavailable as they are difficult to collect and often have ethics issues. Moreover, existing models focus more on learning spatial proximity between trajectories, therefore they can only extract objects with similar spatial features, which are not necessarily traveling companions.

To tackle the aforementioned issues, we develop in this paper an unsupervised %deep
model using autoencoders, which learns the representations directly from original trajectories. The original trajectories are not necessarily aligned along timestamp, and can be of various lengths, i.e., having different numbers of location points. The learned representations contain both spatial and temporal features of the original trajectories, and thereby can be clustered to discover traveling companions.

%\iffalse
%The first part of our model is a LSTM-based autoencoder (LSTM-AE), which takes pre-trained trajectory tokens\footnote{A trajectory token is a vector representation of a location point on the trajectory.} as input and tries to reconstruct the trajectories by minimizing the difference between input and output tokens, encouraging to capture inner correlations between sequential tokens in one trajectory. The pre-trained tokens are embedded with both spatial and temporal information, using skip-gram~\cite{mikolov2013distributed} and positional encoding~\cite{vaswani2017attention}, respectively.
%\fi

The model is inspired by the work of text summarization~\cite{chu2019meansum}, where paragraphs of similar topics are grouped and summarized. Thus we propose to first group the original trajectories using the Sort-Tile-Recursive (STR) algorithm~\cite{leutenegger1997str}. STR is originally used to group spatial data for the bulk-loading construction of R-tree. We use it to group the trajectories so that the trajectories within the same group already have certain spatial and temporal proximity. Then we feed each group of trajectories independently into an attention-based encoder. The idea behind is to encourage closer trajectories to learn more similar representations from each other. Then we use an encoder-decoder structure to reconstruct the input trajectories, and use a decoder-encoder structure to learn the similarity between the input trajectories, as we will show in Section~\ref{model}. The similarity is computed between the mean encodings of the first encoder and the intermediate encodings of the second encoder. As we use global attentions to produce the encoding and a mean operation to aggregate a group of encodings, we call our model ATTN-MEAN Autoencoder. Once we obtain the trajectory encodings using the trained model, we use DBSCAN (as always used in literature~\cite{yao2017trajectory,li2018deep}) to cluster the encodings. The corresponding trajectories in the same cluster are therefore considered as traveling companions.

\section{Related Work}\label{sec:rw}
The problem of finding moving objects that travel together has been extensively studied over the past decade using data mining algorithms. Some representative studies include Flock~\cite{al2007dimensionality}, Convoy~\cite{jeung2008discovery}, Swarm~\cite{li2010swarm} and Gathering~\cite{zheng2013online}. The authors in~\cite{jeung2008discovery} define \textit{convoy} to describe a generic pattern of traveling companions of any shape. A convoy is a group of at least \textit{m} moving objects that are density-connected with respect to a distance \textit{e} during at least \textit{k} consecutive time points. A simple algorithm for discovering convoys is to perform a density-connected cluster algorithm at each time point and maintain convoy candidates who have at least \textit{k} clusters during consecutive time points. Then for each candidate, an intersection of its clusters is conducted to test whether there are at least \textit{m} objects shared by all the clusters. To overcome the high computational complexity, the authors propose to first extract candidate convoys based on simplified trajectories and then decide whether each candidate is indeed qualified in the refinement step. The pattern of \textit{swarms}~\cite{li2010swarm} further relaxes the constraint of convoys by defining traveling companions as moving objects that move within density-connected clusters for at least \textit{k} time points that are possibly non-consecutive. The goal of the work is to discover all \textit{closed swarms}, such that neither the object set nor the number of time points of the swarms can be enlarged. As the number of candidate closed swarms is prohibitively huge, the authors propose two pruning strategies to effectively reduce the search space.

Recently, representation learning of trajectories has drawn lots of attention because it requires very little feature engineering and similarity computation, since using representations naturally avoids the problem of point matching. Studies mostly related to our work include trajectory clustering~\cite{yao2017trajectory} and trajectory similarity computation~\cite{li2018deep,yao2019computing,zhang2019deep} via representation learning. The authors in~\cite{yao2017trajectory} use a seq2seq autoencoder to learn trajectory representations for clustering tasks. They first extract trajectory features such as speed and rate of turn and transform %each trajectory
them into a feature sequence that describes the movement pattern of the corresponding object. Then, they feed the feature sequences into their model to learn fixed-length representations. Later the work in~\cite{li2018deep} proposes an RNN based model to learn trajectory representations for similarity computation. The model does not require feature extraction and directly learns from original trajectories represented by trajectory tokens. However, the model training is supervised and requires to construct training pairs by sampling from the original trajectories. Both studies in~\cite{yao2017trajectory,li2018deep} focus on finding trajectories with similar
% chow2018representation,
shapes, regardless of their actual timestamps. Other studies like~\cite{zhang2019deep} propose to inject additional semantic information such as environmental constraints and trajectory activities into deep models, in order to obtain more accurate trajectory representations for similarity computation. Nonetheless, none of the existing models is designed to learn simultaneously spatial and temporal proximity between trajectories. As such they cannot be directly applied on discovering traveling companions that are both spatially and temporally close.

\iffalse
\section{Problem Definition}\label{sec:problem}
Let $O = \{o_1, o_2, \cdots, o_n\}$ denote a set of $n$ moving objects and  $TR=\{tr_1, tr_2, \cdots, tr_n\}$ denote the corresponding trajectories. Each trajectory $tr$ consists of a sequence of location points $(lp_1, lp_2, \cdots, lp_k)$, where $lp_i$ is a three-dimensional tuple $(t_i, x_i, y_i)$. $t_i$ denotes a timestamp and $(x_i, y_i)$ denote coordinates of $lp_i$. The trajectories may have different number of location points $k$ and do not have to be aligned w.r.t the timestamps $t$.

Then the problem is to find traveling companions whose location points are both spatially and temporally close, that is, they should have a certain number of similar location points $(t, x, y)$. Since the trajectory lengths are varying and the timestamps are not aligned, it is impractical to define similar trajectories based on the three-dimensional tuples. Instead, we propose to learn vectorized representations with fixed length of the trajectories, such that spatially and temporally close trajectories would have similar representations. Then we can discover traveling companions by simply clustering the representations.  
\fi

\section{The models}\label{sec:model}
\iffalse
The fundamental part of our model is the pre-training of trajectory points. we divide the space where trajectory points scatter into small cells and use skip-gram to train the representation of cells by learning from relationship between cells and their surrounding ones. Positional encoding has been used here to embed temporal information. The main body of the model is composed of two tied autoencoders. One is used to capture trajectories' inner correlations by reconstructing inputs. The other is used to capture outer correlations between trajectories and its closer trajectories differed by Sort-Tile-Recursive algorithm.
\fi

\subsection{Representation of Trajectory Points}\label{sec:lstmae}
Firstly, we divide the entire space into square cells of the same size. Then, following the idea in~\cite{li2018deep}, we use the skip-gram technique~\cite{mikolov2013distributed} to pre-train the cell representations (or token) such that spatially close cells have similar representations. So, each location point of a trajectory can be represented using the token of the corresponding cell the point falls in.

% \iffalse
% The first problem is to convert the original representation of a trajectory into a ordered sequence of fixed-length vectorized tokens.
% % , where each token is a fixed-length vector representing a location point in the original trajectory. 
% % The representations should reflect both spatial and temporal proximity between the original location points. 
% In order to obtain a representation reflecting both spatial and temporal proximity between points within trajectories, we leverage two techniques to embed the most important spatial and temporal features into trajectory tokens.
% \fi

% \iffalse
% The second issue is to capture temporal information of each location point. A naive approach is to view the entire spatial-temporal space as a three-dimensional space, and grid this space into three-dimensional cells. Then again one can use skip-gram to train the cell representations as what is done for the two-dimensional cells. However, this approach produces a huge number of cells, making it intractable to train. 
% \fi
% ~\cite{vaswani2017attention}
Secondly, to capture temporal characteristics, we propose to use the positional encoding~\cite{gehring2017convolutional} technique to inject time information into the above trajectory tokens. Positional encodings have been used in the self-attention model to capture information about token positions in a sequence. As a trajectory is an ordered sequence of tokens, we could use the same mechanism to model the sequence. We compute the positional encodings as Equation \ref{pos} and \ref{pos_2}:
 	\begin{equation}\label{pos}
 		PE(t_{i}, 2p) = sin(\frac{t_{i}}{10000^{{2p}/{d_{cell}}}})
	\end{equation}
 	\begin{equation}\label{pos_2}
        PE(t_{i}, 2p + 1) = cos(\frac{t_{i}}{10000^{{2p}/{d_{cell}}}})
	\end{equation}
where $t_{i}$ is the $i^{th}$ timestamp of the entire time duration in the data, $d_{cell}$ indicates the embedding size of a cell, and $p \in [0, d_{cell} / 2 - 1]$. Each positional encoding has the same size with the above token, and is therefore added directly to it.

\subsection{ATTN-MEAN Autoencoder}
\iffalse
The first autoencoder of the model is a simple LSTM based autoencoder (LSTM-AE). In the process of encoding, the encoder reads the input trajectory tokens step by step. After reading in the entire sequence, we leverage global attention mechanism to aggregate trajectory encodings into fixed a fixed-length vector, which represents a learned representation of the original trajectory. Then, in the process of decoding, the aggregated fixed-length hidden state works as initial state of the LSTM decoder, and the decoder interprets encodings from encoder and generate output tokens to reconstruct a lossy input trajectory. Once the reconstruction loss converges, we stop training the autoencoder and use the encoding generated by the encoder as trajectory representations. The reconstruction loss is computed using mean square error (MSE), since using cross entropy here will lose temporal information, as tokens with the same location, but with different timestamps have different meaning.
\fi

In order to encourage trajectories to learn more from their neighbours, we propose to first roughly divide them into groups w.r.t their spatial-temporal proximity, and then feed each group independently into the other autoencoder. We use the Sort-Tile-Recursive (STR) algorithm~\cite{leutenegger1997str} to group the trajectories. The STR algorithm is originally used to pack spatially-close objects into minimum bounding rectangles (MBRs) and construct a bulk-loading R-tree index of the objects. Since each trajectory can be considered as an object in a three-dimensional space $(t, x, y)$, where $t$ is the time dimension and $(x, y)$ stands for the two spatial dimensions, we apply the idea of STR to pack trajectories. Trajectories in the same MBR or group are more likely to be traveling companions, and thus fed as an independent group into the model.

\begin{figure}
\includegraphics[width=0.5\textwidth]{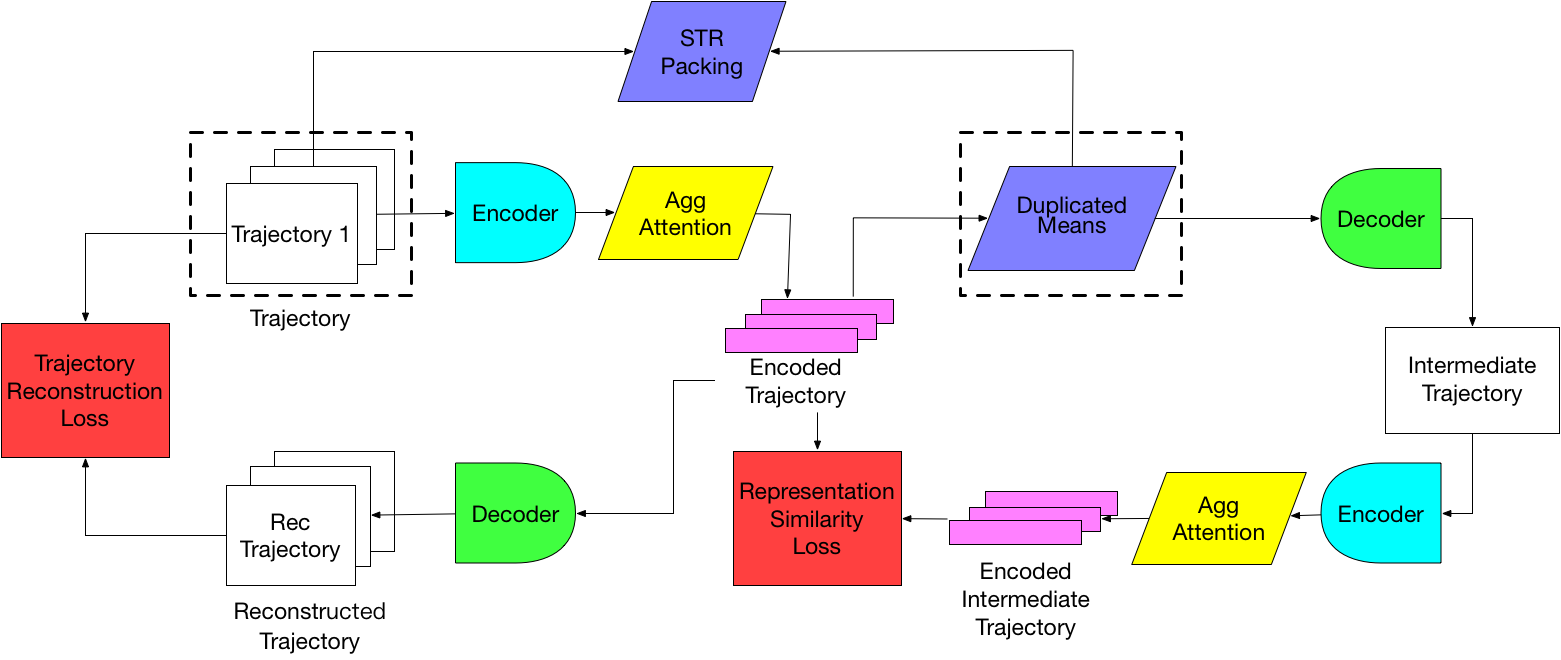}
\caption{The proposed ATTN-MEAN model} \label{model}
\end{figure}

% Inspired by the work in text summarization~\citep{chu2019meansum}, 
The entire model is constructed using two autoencoders, as depicted in Figure~\ref{model}. The encoder and decoder on the left-hand side constitute the first autoencoder, which is an ordinary LSTM autoencoder used to reconstruct the trajectories. On the right-hand side, the encodings are firstly averaged using a mean operation and then fed into a decoder. The decoded intermediate trajectories are again input into an encoder to obtain the encoded intermediate mean trajectories. Then we force the encodings output by the left encoder to learn from each other by computing the similarity between them and the encodings produced by the right encoder. Thus the autoencoder on the right-hand side is indeed a decoder-encoder structure. The two encoders and the two decoders have the same structures and share the same parameters, respectively.

\iffalse
We train the entire two encoder-decoder components jointly, but with two types of loss functions, namely, \emph{trajectory reconstruction loss} and \emph{representation similarity loss}. The two encoders and the two decoders share the same weights. The complete architecture is presented in Figure~\ref{model} and the details are explained below.
\fi

On the two LSTM encoders, we additionally use a global attention mechanism to aggregate the hidden states of each step to form the output encodings. The encoding contains more overall information of the entire trajectory and boosts the model's performance in the experiments. In particular, we firstly initialize a global attention vector $a$ to calculate an attention score on each trajectory token, as shown in equation~\ref{attentionweight},
\begin{equation}\label{attentionweight}
	 \alpha_i = \frac{\exp(h_i^T \cdot a)}{\sum_{j=1}^{l_{tr}}\exp{h_j^T \cdot a}}
\end{equation},
where $h_i$ is the hidden representation of point $i$, $\alpha_i$ is the attention weight of this representation and $l_{Tr}$ is the length of each trajectory. Then, we use these attention scores to conduct a weighted sum of the hidden state vectors of each trajectory, denoted by $H_{tr}$,
\begin{equation}\label{trajvector}
	 H_{tr} = \sum_{i=1}^{l_{tr}}\alpha_i \cdot h_i
\end{equation}
, where $H_{tr}$ is the final representation of the trajectory.

For optimization, we compute the trajectory reconstruction loss $l_{rec}$ using mean squared error (MSE) for the left component, as shown in Equation~\ref{rec},
\begin{equation}\label{rec}
    \iota_{rec} (\{(tr_1, {tr_1}'), \cdots, (tr_k,{tr_k}')\}) = \sum_{i=1}^{k}l_{MSE}(tr_i, {tr_i}')  
\end{equation}
where $k$ is the batch size, $tr_i$ denotes the $i^{th}$ trajectory and ${tr_i}'$ is the corresponding reconstructed trajectory. Then we compute the representation similarity loss between the trajectory encodings produced by the two encoders using average cosine distance, as shown in Equation~\ref{sim},
 \begin{equation}\label{sim}
    \iota_{sim} (\{(tr_{1}, tr'), \cdots, (tr_{k}, tr'\}) = \frac{1}{k} \sum_{i=1}^{k} d_{COS}(H_{tr_{i}}, H_{tr}')  
\end{equation}
where $H_{tr_{i}}$ denotes the aggregated encoding of trajectory $tr_{i}$ generated by the encoder in the left component and $H_{tr'}$ denotes the intermediate trajectory encoding generated by the encoder in the right component.

Through minimizing both the reconstruction loss and similarity loss, we force the encodings produced by the encoder to keep the distinctive features of their own original trajectories as well as learn similar features from their neighbours in the same group. As we use global attentions and a mean operation inside and outside the encoder, we call this model ATTN-MEAN Autoencoder.

\section{Performance Evaluation}\label{sec:exp}
% We study the effectiveness and superiority of our proposed method on two real-world datasets. The experimental setup, parameter setting and training details are presented in ~\ref{sec:expsetup} and ~\ref{sec:paradetail} separately.
\subsection{The Dataset and Overall Settings}\label{sec:expsetup}
We obtain a trajectory dataset of the passengers in our collaboration with an airport in Asia. The dataset contains 14605 trajectories with about 719,507 points, and each trajectory has 20 to 120 location points. We attempt to use this data to find passengers who walk together.

Using the dataset, we compare ATTN-MEAN with an LSTM autoencoder (LSTM-AE), Convoy~\cite{jeung2008discovery}, Swarm~\cite{li2010swarm}, T2VEC~\cite{li2018deep}, BFEA~\cite{yao2017trajectory}.
For two pattern mining based methods Convoy and Swarm, we compare the number of extracted clusters with that discovered using our learned representations. To comply with the two algorithms, we conduct linear interpolations in the dataset, such that we generate a synthetic point every 10 seconds for each trajectory, if necessary. For T2VEC and BFEA, we also inject positional encodings into their original input for a fair comparison. We use DBSCAN to cluster the learned encodings and compare the clustering performance. We only list the main results due to page limit. More experimental results are available upon request.

\subsection{Parameter Settings and Training Details}\label{sec:paradetail}
\textbf{Group size and batch size}. We set the group size and batch size to 64 and 8, respectively. The group size is the size of MBR capacity used in the STR algorithm.

\textbf{Cell Size.} We divide the entire airport space into square cells with length 5 meters on each side, and obtain $15,354$ cells.
% This is a relatively small vocabulary size and thus the pre-training of cell representations using skip-gram is quite efficient.

\textbf{Embedding Size.} We set the embedding size of the cell representations to be 256. Therefore the positional encoding for each timestamp also has 256 dimensions.

% ~\cite{kingma2014adam}
\textbf{Training Details.} We use Adam stochastic optimization for training with the learning rate $0.001$ and weight decay rate $0.00001$. The training process is terminated when trajectory reconstruction loss and representation similarity loss both converge. We observe that all the three models converge after 20 epochs.

\subsection{Main Results}
\begin{figure}
\centering
\subfigure[Davies Bouldin Index]{
\begin{minipage}[t]{0.3\linewidth}
\centering
\includegraphics[scale=0.12]{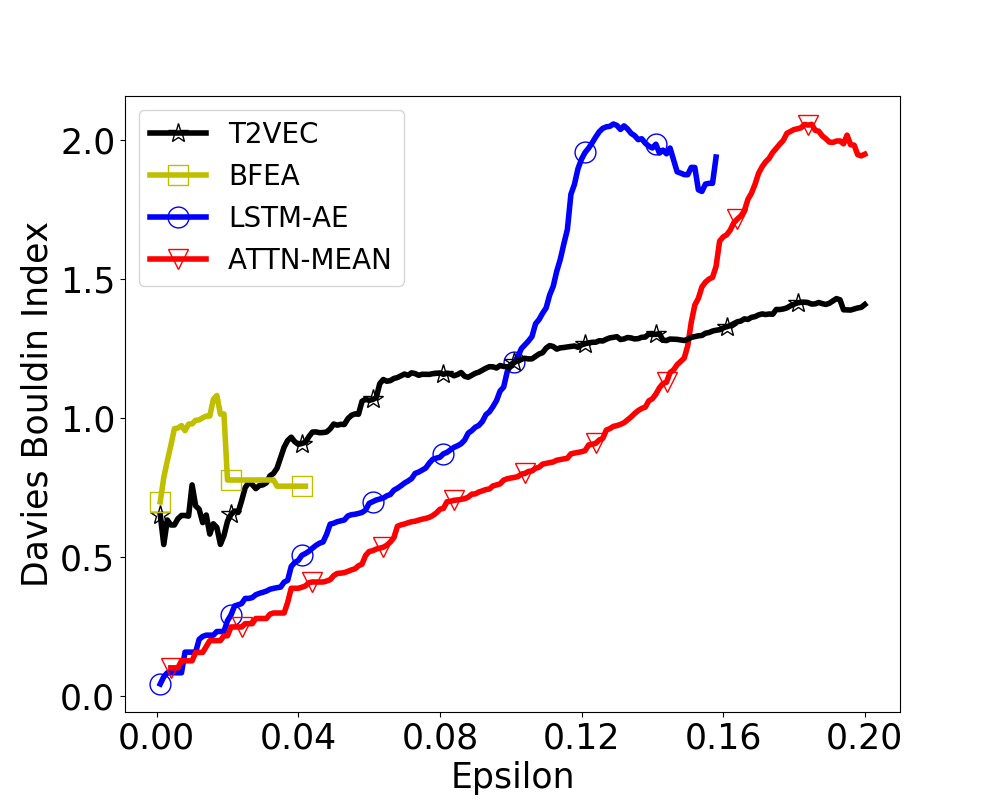}
\label{davies}
\end{minipage}
}
\subfigure[Silhouette Coefficient]{
\begin{minipage}[t]{0.3\linewidth}
\centering
\includegraphics[scale=0.12]{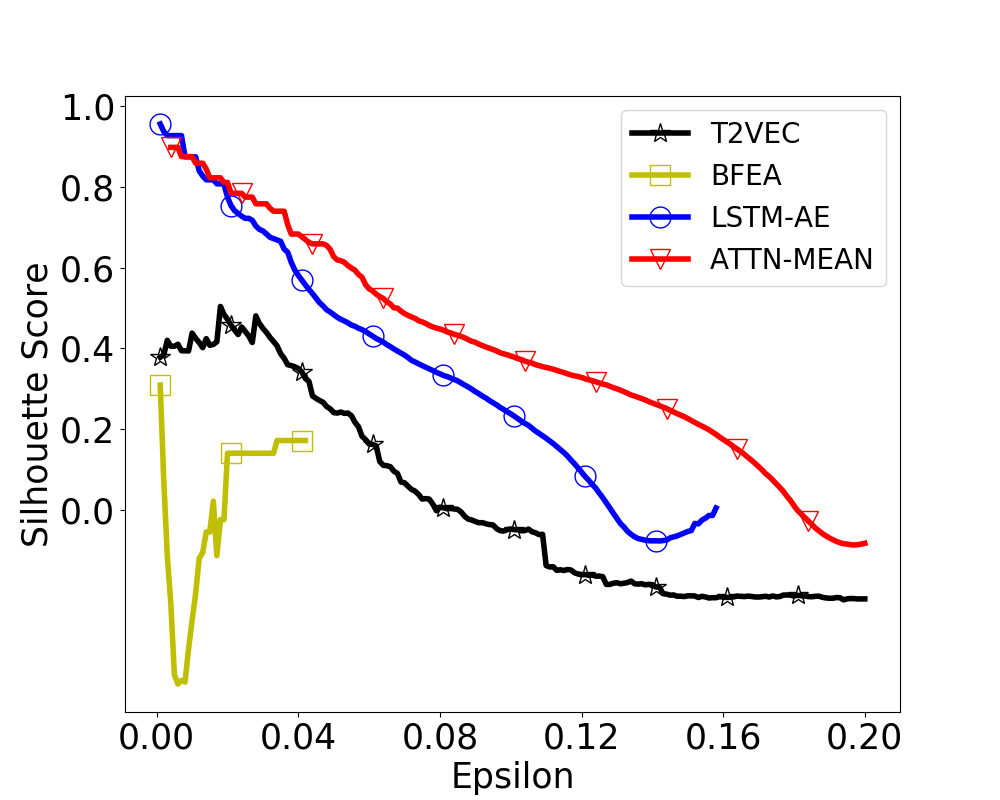}
\label{silscore}
\end{minipage}
}
\subfigure[Weighted Average Entropy]{
\begin{minipage}[t]{0.3\linewidth} 
\centering
\includegraphics[scale=0.12]{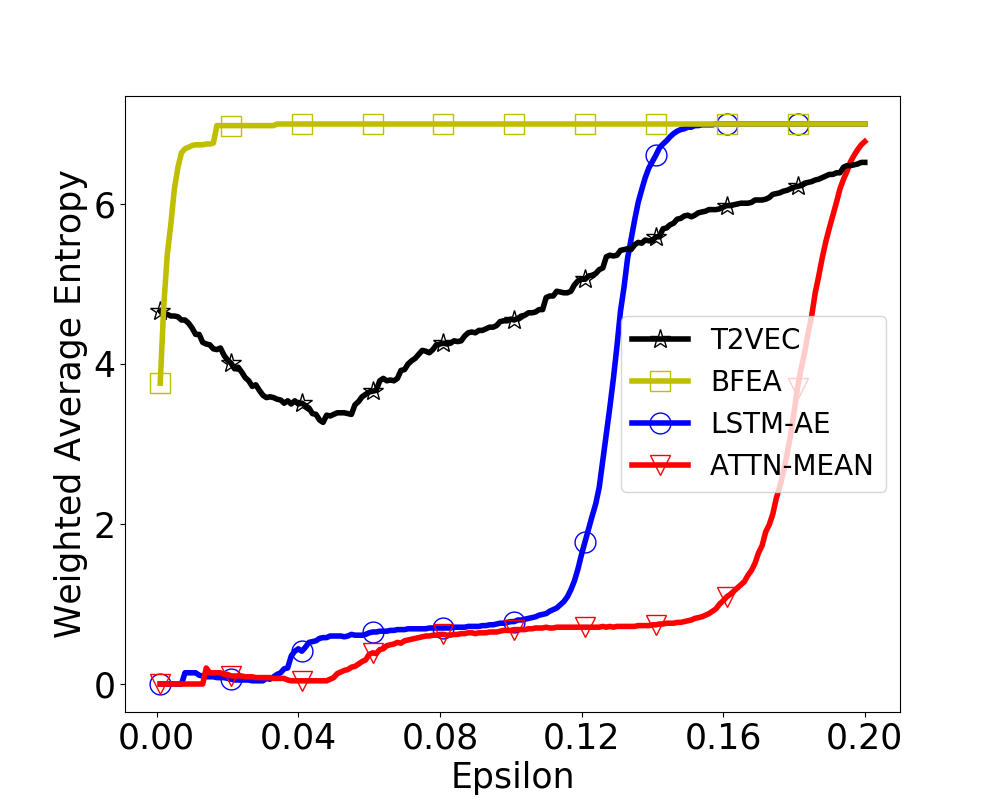}
\label{eps-entropy}
\end{minipage}
}
\centering
\caption{Clustering evaluation using three metrics.}
\end{figure}

We employ three metrics to evaluate the clustering results, namely, Davies-Bouldin Index, Silhouette Coefficient and weighted average entropy. A smaller Davies-Bouldin index and a high Silhouette Coefficient value indicate better clustering performance in general. To calculate the weighted average entropy of each cluster, we use the nearest gate of each trajectory's last point as its label. The idea behind is that traveling companions are more likely to walk to the same gate. For all these measurements, we discard all clusters of size one, i.e., trajectories that are considered as traveling alone. We vary the distance parameter $epsilon$ of DBSCAN from 0 to 0.2 with increments $0.0001$ and plot the three metrics for each $epsilon$. The results are presented in Figure~\ref{davies}, ~\ref{silscore} and ~\ref{eps-entropy}, respectively.

We observe in Figure~\ref{davies} and~\ref{silscore} that in general ATTN-MEAN has smaller Davies-Bouldin Index and larger Silhouette Coefficient for varying $epsilon$, suggesting that ATTN-MEAN performs better than LSTM-AE, T2VEC and BFEA for both internal and external criteria evaluation.
We also observe in Figure~\ref{eps-entropy} that ATTN-MEAN produces smaller weighted average entropy than LSTM-AE, T2VEC and BFEA for varying $epsilon$. This means that clusters produced by ATTN-MEAN have fewer different gate labels, which means the clustered trajectories are more likely to be travelling companion.
% , thanks to the packing strategy and multi-head attention in the representation learning. 

% Meanwhile, we can tell that autoencoder-based model performs better than other models on most situations, which also demonstrate that autoencoder-based methods are better than others. 
Besides, We compare the number of clusters that have at least two trajectories found by the two models with that extracted by Convoy and Swarm. The results are presented in Table 1. For Convoy and Swarm, we set $k$ to be 18 (i.e., at least 3 minutes), $m$ to be 2, and vary $e$ in $(3m, 5m)$. For example, the first row in the table means that a convoy should have at least two trajectories with distance less than 3 meters in at least 18 consecutive time points, where two consecutive time points have 10-second offset. For our two models, we simply show the results when they discover the largest number of clusters. 
\begin{table}[h]
\centering
 \resizebox{0.46\textwidth}{12mm}{
\begin{tabular}{|c|c|c|c|}
\hline
Algorithm & Parameters&Number of Clusters & Single trajectories\\
\hline
\multirow{2}{*}{Convoy}&k=18,m=2,e=3$m$&308&4629\\
\cline{2-4}
&k=18,m=2,e=5$m$&814&3875\\
\hline
\multirow{2}{*}{Swarm}&k=18,m=2,e=3$m$&756&4086\\
\cline{2-4}
&k=18,m=2,e=5$m$&2431&2710\\
\hline
LSTM-AE&&786&895\\
\hline
ATTN-MEAN&&727&1112\\
\hline
\end{tabular}}
\caption{The clusters discovered by different algorithms.}
\end{table}
We observe that even for these relaxed parameter settings, Convoy and Swarm generate lots of single trajectories that do not belong to any cluster. By contrast, our models can group most trajectories into clusters.

\subsection{The Effect of Positional Encodings}
% Positional encodings are used to efficiently capture temporal information in the trajectory token representation. To show the effectiveness of this idea, we remove only the positional encodings from the token representations and train the ATTN-MEAN model. We compute the weighted average entropy and the total number of clusters for varying $epsilon$. The results are presented in Figure~\ref{posentropy} and~\ref{posnum}.

\begin{figure}
	\centering
	\subfigure[Davies Bouldin Index]{
	\begin{minipage}[t]{0.3\linewidth}
	\centering
	\includegraphics[scale=0.12]{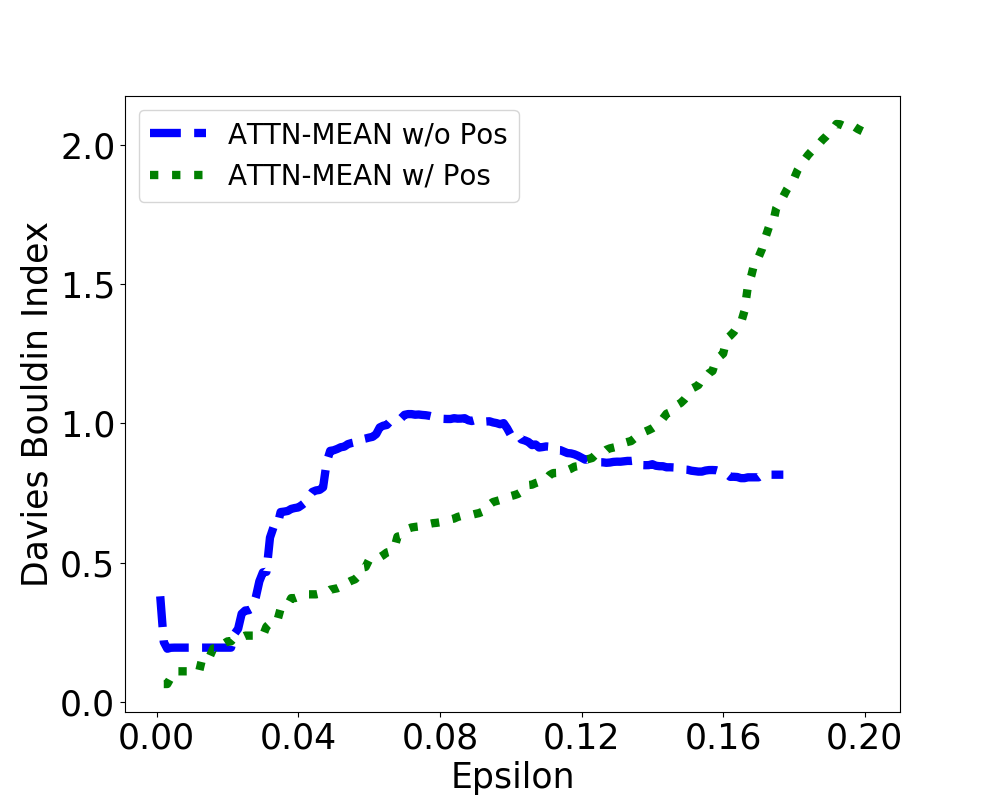}
% 	\label{posnum}
	\end{minipage}
	}
	\subfigure[Silhouette Coefficient]{
	\begin{minipage}[t]{0.3\linewidth}
	\centering
	\includegraphics[scale=0.12]{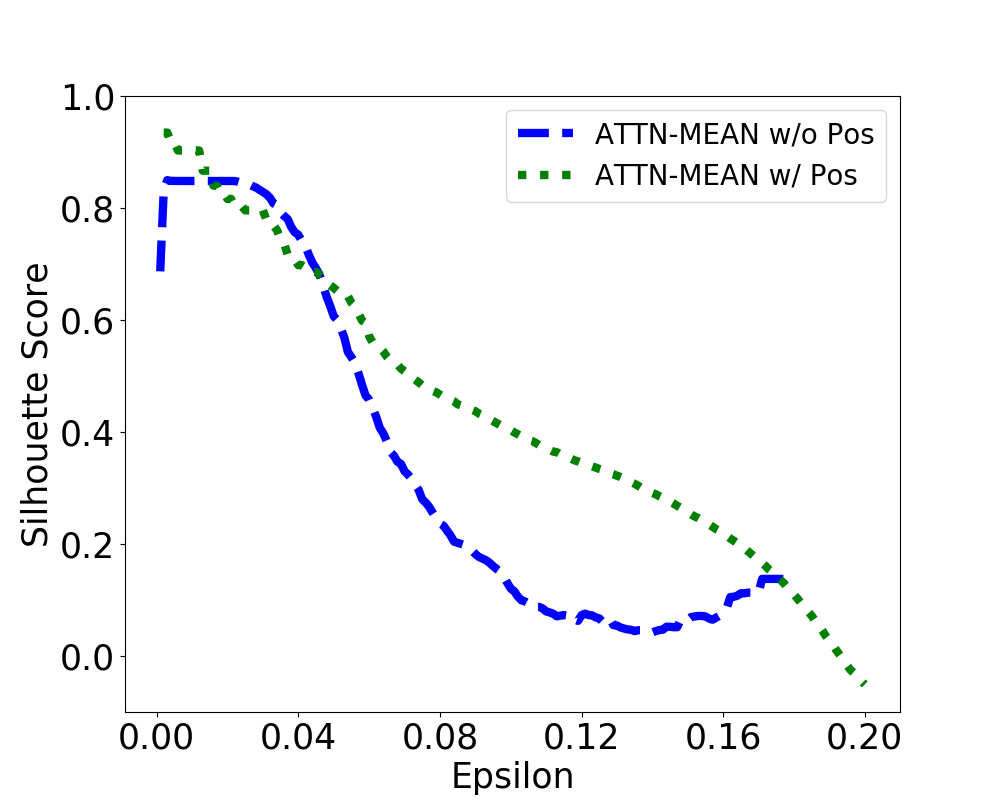}
% 	\label{posnum}
	\end{minipage}
	}
	\subfigure[Weighted average entropy]{
	\begin{minipage}[t]{0.3\linewidth}
	\centering
	\includegraphics[scale=0.12]{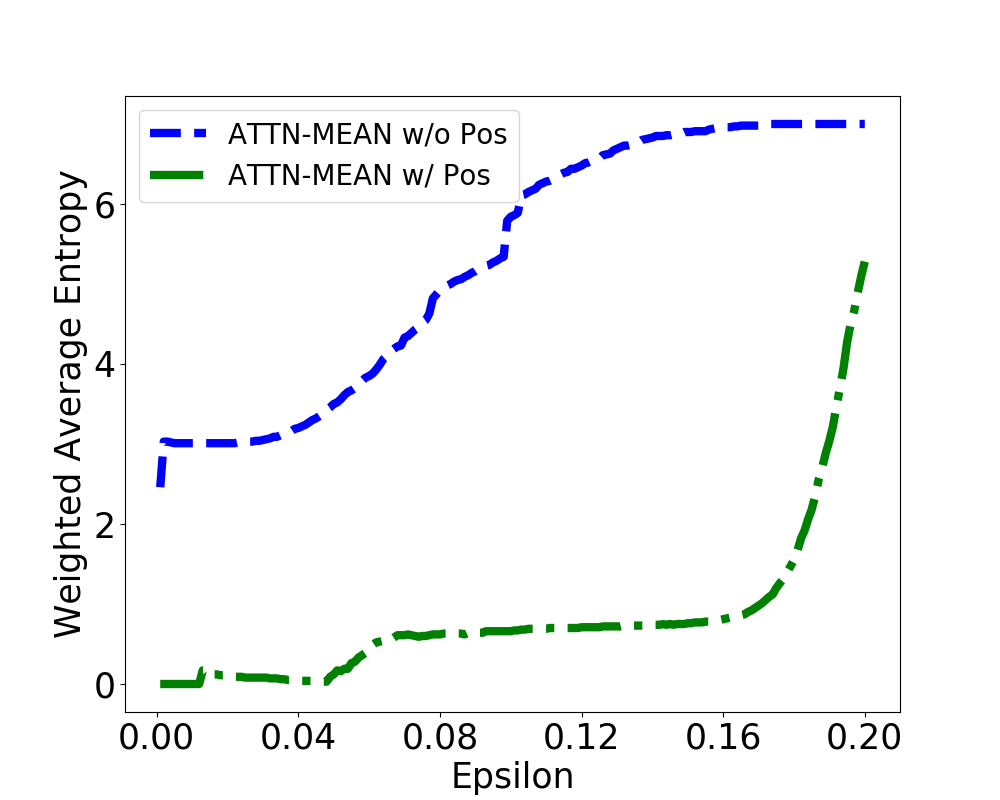}
	\end{minipage}
	}
	\centering
	\caption{The Effect of Positional Encodings.}
	\label{posentropy}
\end{figure}

We also conduct an ablation study to show the effect of positional encodings. We observe in Figure~\ref{posentropy}, ATTN-MEAN performs generally better than ATTN-MEAN without positional encodings for varying $epsilon$ on all the three metrics. This proves that the injection of positional encodings helps our model to find trajectories that are both spatially and temporally close. The model without positional encodings would find trajectories with similar shape in different time periods.
% we can tell positional encoding is very important in enriching the trajectory representations, capturing both spatially and temporally close trajectories.
% Without positional encodings, ATTN-MEAN can only learn spatial proximity between trajectories. In this case, trajectories with similar shapes but deviating on the time dimension are still likely to have similar representations, thereby being misidentified as traveling companions. 

% many trajectories of different periods have similar trajectory shapes
% This is also illustrated by the total numbers of clusters shown in Figure~\ref{posnum}. When $epsilon$ is very small, ATTN-MEAN without positional encodings already produces more than 1500 clusters. This is because many trajectories of different periods have similar trajectory shapes, especially in the airport terminal. Thus there are much more similar learned representations compared to those learned by ATTN-MEAN with positional encodings, which form many clusters when the distance parameter is still small. By contrast, ATTN-MEAN with positional encodings gradually finds more clusters as $epsilon$ increases. Then the number of clusters decreases as more and more trajectories are clustered together, until all trajectories are grouped into one cluster.

\section{Conclusion and Future Work}\label{sec:conclusion}
In this work, we propose an unsupervised deep representation model ATTN-MEAN for the discovery of traveling companions.
% The  ATTN-MEAN is composed of a simpe LSTM-based autoencoder, where 
We first employ positional encoding and skip-gram techniques to embedding the trajectories. The input trajectory token representations are collectively embedded with spatial and temporal information of the location points. Then we use STR to group original trajectories to encourage them learn from the neighbours. A double autoencoder architecture with global attentions and a mean operation is used to construct the model. Experimental results show that ATTN-MEAN learns overall better trajectory representations than LSTM-AE, T2VEC and BFEA for discovering traveling companions. In future, we plan to explore other mechanisms to fuse the two autoencoders, and further improve ATTN-MEAN.

% In future, we plan to study various detailed architectures of the autoencoders such as self-attention and memory-augmented networks, to further improve the representation learning of the trajectories.

\bibliographystyle{ACM-Reference-Format}
\bibliography{reference}

\end{document}